\documentclass[runningheads]{llncs}
\usepackage{cite}
\usepackage[pdftex]{graphicx}
\usepackage{epstopdf}
\usepackage{algorithmic}
\usepackage[linesnumbered,ruled,vlined]{algorithm2e}
\usepackage{subcaption}
\usepackage{multirow}
\usepackage{indentfirst}
\usepackage{amsmath}
\usepackage{float}
\usepackage[misc]{ifsym}
\usepackage{makecell}

\begin{document}
\title{Improving Question Answering over Knowledge Graphs Using Graph Summarization\footnote{The paper has been accepted by ICONIP 2021.}}
\titlerunning{GS-KGQA}
%
\author{Sirui Li\inst{1} \and
Kok Wai Wong\inst{1} \and
Chun Che Fung\inst{1}\and
Dengya Zhu\inst{1,2}}
\authorrunning{Sirui.Li et al.}
 \titlerunning{GS-KGQA}
\institute{Discipline of Information Technology, Murdoch University, South St, Murdoch, Western Australia \and
Discipline of Business Information Systems, School of Management and Marketing, Curtin University, Kent St, Bentley, Western Australia\\
\email{}}
\toctitle{Improving Question Answering over Knowledge Graphs Using Graph Summarization}
\tocauthor{Sirui~Li; Kok Wai~Wong; Chun Che~Fung; Dengya~Zhu}
\maketitle              
\begin{abstract}
 Question Answering (QA) systems over Knowledge Graphs (KGs) (KGQA) automatically answer natural language questions using triples contained in a KG. The key idea is to represent questions and entities of a KG as low-dimensional embeddings. Previous KGQAs have attempted to represent entities using Knowledge Graph Embedding (KGE) and Deep Learning (DL) methods. However, KGEs are too shallow to capture the expressive features and DL methods  process each triple independently. Recently, Graph Convolutional Network (GCN) has shown to be excellent in providing entity embeddings. However, using GCNs to KGQAs is inefficient because GCNs treat all relations equally when aggregating neighbourhoods. 
Also, a problem could occur when using previous KGQAs: in most cases, questions often have an uncertain number of answers. To address the above issues, we propose a graph summarization technique using Recurrent Convolutional Neural Network (RCNN) and GCN. The combination of GCN and RCNN ensures that the embeddings are propagated together with the relations relevant to the question, and thus better answers. The proposed graph summarization technique can be used to tackle the issue that KGQAs cannot answer questions with an uncertain number of answers. In this paper, we demonstrated the proposed technique on the most common type of questions, which is single-relation questions. Experiments have demonstrated that the proposed graph summarization technique using RCNN and GCN can provide better results when compared to the GCN. The proposed graph summarization technique significantly improves the recall of actual answers when the questions have an uncertain number of answers.
\end{abstract}
\section{Introduction}
Question Answering (QA) is a traditional task in Natural Language Processing (NLP). In order for the QA system to work effectively, it is necessary to precisely understand the meaning of the natural language question and select the most appropriate answers from various background information. There are  two main branches in the QA system: QA over Knowledge Graphs (KGs) (KGQA) and QA over documents. Compared to QA over documents, KGQA could provide more comprehensive answers: for a given question, the answer may come from various resources in the KGQA~\cite{deng2018deep}. Thereby, KGQA becomes an important topic and attracts much attention recently~\cite{huang2019knowledge, saxena2020improving,zhang2020answer, de2018question, sun2019pullnet, sun2018open}. 
A KG is a directed graph with real-world entities as nodes and their relations as edges~\cite{huang2019knowledge}. In the KG, each directed edge, along with its head entity and tail entity, constitutes a triple, i.e., (head entity, relation, tail entity).

The key idea of recent KGQAs is to represent questions and entities/relations as low-dimensional vectors or embeddings~\cite{huang2019knowledge}. Knowledge Graph Embedding (KGE) aims to learn a low-dimensional vector for each relation/entity in a KG, such that the original relation features are well preserved in the vectors~\cite{huang2019knowledge}. These learned vectors have been employed to complete many KGQAs efficiently because they can predict missing links between entities~\cite{huang2019knowledge, saxena2020improving, li2021question}. However, these shallow KGEs are limited to their expressiveness~\cite{xie2020reinceptione}. Therefore, some KGQAs~\cite{qiu2020stepwise, miller2016key,li2021question} have used deep learning methods, such as Long Short-Term Memory (LSTM), to represent relations/entities in a KG. Nevertheless, all these KGQAs process each triple independently, i.e., they cannot capture semantically rich neighbourhoods and hence produce low-quality embeddings~\cite{arora2020survey}.

Recently, Graph Convolutional Network (GCN) has been widely adopted for graph embedding due to its excellent performance and high interpretability~\cite{kipf2016semi}. GCN is a multi-layer neural network operating on graph-structural data. GCN models relevant information about node's neighbourhoods as a low-dimensional feature vector~\cite{kipf2016semi}. Unlike KGE or other deep learning methods, GCN well preserves the structural relation features and semantic neighbour features in the vectors~\cite{li2021learning}. As reported in the literature, the effectiveness of GCN is the main motivation for this paper to investigate its usage as the graph embedding technique in KGQAs.

Applying GCNs to KGQAs has three significant challenges. Firstly, simple GCNs are inefficient in KGQAs because GCNs treat all relations equally when aggregating neighbourhoods for each target node~\cite{arora2020survey}. However, in most cases, it should be anticipated that the entity embeddings are propagated more along with the relations relevant to the question to provide a more relevant answer. Secondly, a relation often has various expressions in natural language questions, posing an additional challenge in question analysis~\cite{huang2019knowledge}. For instance, the relation ``person.nationality" can be expressed as “what is ... ’s nationality”, “which country is ... from”, “where is ... from”, etc. Thirdly, most current KGQAs cannot effectively answer questions with multiple entities or when there is an uncertain number of answers. They assume that a question only has one entity and could be answered by a single entity~\cite{aghaebrahimian2017hybrid}.  Nevertheless, a question could be \textit{``Which films are co-acted by Vin Diesel and Paul Walker?"}. It requires the KGQA to identify two entities and return a series of films. Answering such questions requires more knowledge of the relations~\cite{aghaebrahimian2017hybrid}. Some GCN-based KGQAs~\cite{sun2018open, sun2019pullnet} answer such questions using a binary classifier, which sometimes could miss some correct answers or predict additional incorrect answers. The most popular solution is to predict answers based on the \textit{softmax} function~\cite{de2018question, zhang2020answer, huang2019knowledge}. The softmax function assigns scores to all candidates, and then KGQAs select candidates whose scores are close to the highest score within a certain margin. The challenge here is to set a proper margin for every question because the number of answers could be uncertain.

Through analyzing the above three challenges, we propose a Graph Summarization based KGQA (GS-KGQA). GS-KGQA combines GCN with Recurrent Convolutional Neural Network (RCNN) to solve the first two challenges. RCNN has already been successfully used in text classification~\cite{lai2015recurrent}. In our work, RCNN is used to classify a given question, and to predict the probabilities of all relations. These predicted probabilities are the relational weights used in the variant of the GCN used in this paper. This combination ensures that embeddings are propagated more along with edges relevant to the question, i.e., by propagating relations with high probabilities to ensure that the answer provided is the most relevant. The RCNN also converts the relation analysis of questions to the classification task. We propose two graph summarization algorithms to solve the third challenge; one used for questions with a single entity and another used for questions with multiple entities. Our graph summarization algorithms group candidate answers into one node prior to the softmax function.

In this paper, we demonstrated the proposed graph summarization algorithms on the most common type of questions, which is single-relation questions~\cite{huang2019knowledge}. Single-relation questions contain only one relation. In summary, the contributions in this paper are:
\begin{enumerate}
\item The proposed GS-KGQA system combines GCN and RCNN to ensure that the embeddings can propagate along with the desire relations relevant to the question.
    \item The proposed GS-KGQA system with the proposed graph summarization algorithms are investigated for the single-relation KGQA task and to address the issue of an uncertain number of answers. This is further investigated for single-relation questions with multiple entities.
    \item Assess the effectiveness of GS-KGQA over baselines using four benchmark datasets.

\end{enumerate}
\section{Related Works}
Among many recent deep learning tasks~\cite{liu2021invertible, zhu2017evaluation, goh2006towards,li2020transferring, li2020tspnet}, KGQA is one of the most important area. 
A series of work~\cite{huang2019knowledge, saxena2020improving, li2021question} have been performed in KGE to learn the low-dimensional representations of entities and relations in a KG as follows. Huang et al.~\cite{huang2019knowledge} jointly recovered the question’s head entity, relation, and tail entity representations in the KGE spaces. Saxena et al.~\cite{saxena2020improving} took KGQA as a link prediction task and incorporated ComplEx, a KGE method, to help predict the answer. However, these shallow KGE models are limited to their expressiveness~\cite{xie2020reinceptione}. Later, deep learning embedding-based KGQAs~\cite{qiu2020stepwise, miller2016key,li2021question} have been proposed to capture the expressive features with efficient parameter operators.
Yunqiu et al.~\cite{qiu2020stepwise} utilized a bidirectional Gated Recurrent Unit (GRU) to project questions and relations/entities into the same space. 
However, they process each triple independently and produce low-quality embeddings~\cite{arora2020survey}.

Many GCN-based KGQAs have been proposed to capture both semantic features and structural features in entity embeddings~\cite{de2018question, sun2019pullnet, sun2018open, zhang2020answer}. Nicola et al.~\cite{de2018question} explored the structural features of GCN to find missing links between entities. Instead of embedding the whole knowledge graph, Sun et al.~\cite{sun2018open, sun2019pullnet} extracted question-related subgraphs and then updated node embeddings by a single-layer GCN. They selected answers based on binary classifiers. Zhang et al.~\cite{zhang2020answer} integrated question information into the subgraph node embedding to obtain the node presentation with question dependency. However, simple GCNs ignore the edge labels in the graph~\cite{arora2020survey}. Few efforts have been given to the questions with an uncertain number of answers. It is therefore one of the objectives of this paper to investigate this challenge.
\section{GS-KGQA: Proposed Method}
\subsection{GS-KGQA Overview}
\begin{figure}
\centerline{\includegraphics[scale=0.35]{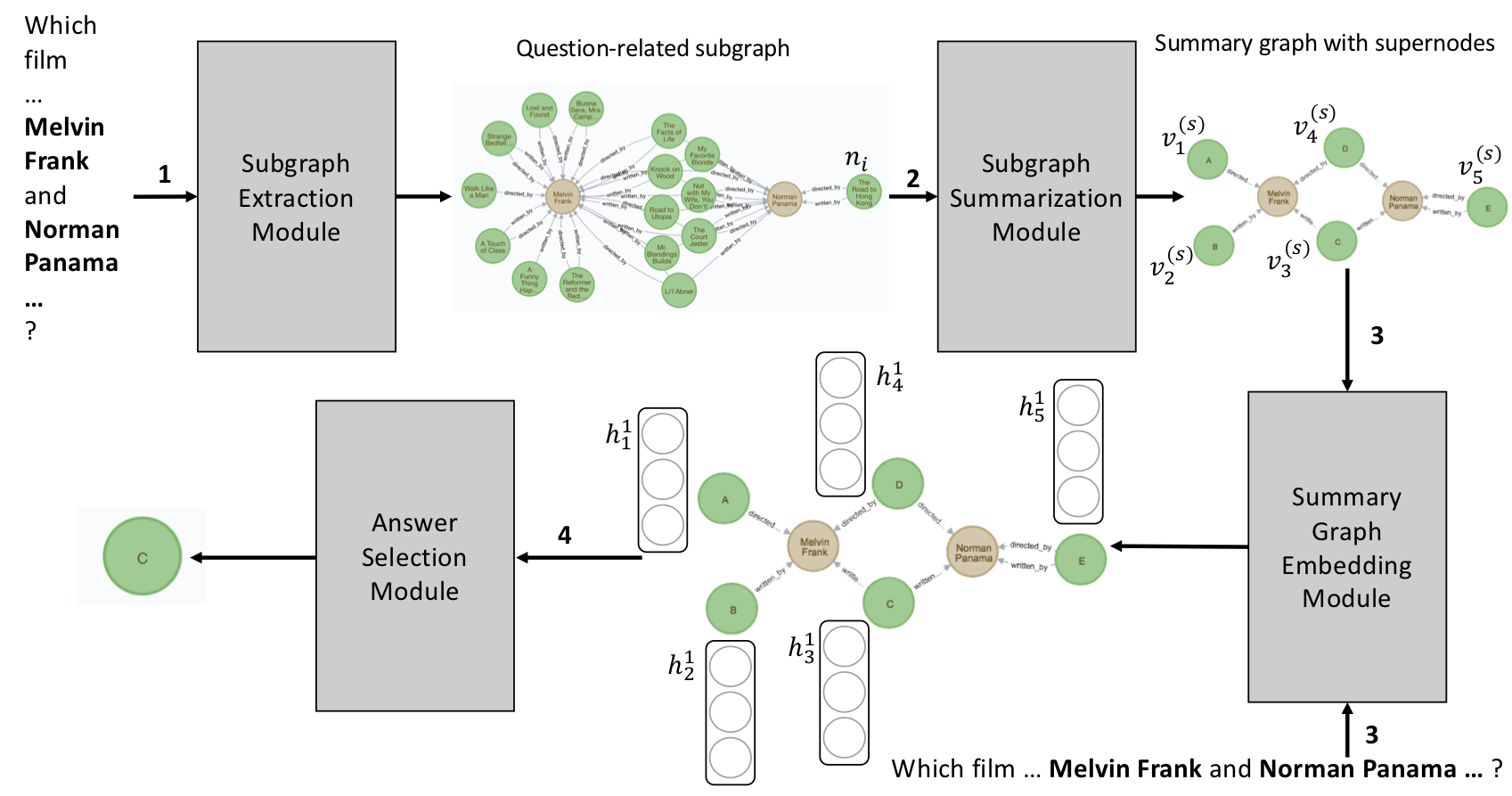}}
\caption{The overview of GS-KGQA}
\label{fig:system}
\end{figure}
The framework of the proposed GS-KGQA is illustrated in Figure~\ref{fig:system}. It consists of four modules. They are introduced from Section 3.2 to Section 3.5.
\subsection{Subgraph Extraction Module}
This module finds the question-related subgraph from the KG. It identifies which entities are mentioned in a given question and links the entities to the KG. Named Entity Recognition (NER) identifies mentions in the text. After comparing some NER methods, we selected spaCy\footnote{https://spacy.io/} for its quick and straightforward implementation. 
Then, all triples containing the identified entities are extracted from the movie KG to form the question-related subgraph. Note that this module could be replaced by any advanced NER methods.
\subsection{Subgraph Summarization Module}
The subgraph summarization module groups the question-related subgraph to a summary graph. In this paper, we highlight that single-relation questions could involve multiple entities. Different graph summarization algorithms are proposed for questions with a single entity and with multiple entities, respectively. Note that we call the entity mentioned in a given question as \textit{question entity}; the node directly linked with a node is called \textit{neighbour}.

The algorithm for questions with a single entity is presented in Algorithm~\ref{alg: one-hop-based}. Neighbours linked with the question entity by the same relation are aggregated into one supernode. Note that one neighbour might be grouped into different supernodes. 
\begin{algorithm}
\SetAlgoLined
\KwIn{A question-related subgraph $G=<V, E, R>$ from the subgraph extraction module; the question entity $v_c$. $V$ is the node set, $E$ is the edge set and $R$ is the relation set.}
\KwOut{A summary graph $G_s = <V_s, E_s, R>$}
\For{$r_i$ in $R$}{
$supernode$ = set()\;
\For{neighbour $v_i$ of $v_c$}{
\If{$v_i$ and $v_c$ is linked by $r_i$}{
$supernode$.add($v_i$)
}
}
$V_s$.add($supernode$)
}
 \caption{Algorithm for questions with a single entity}
 \label{alg: one-hop-based}
\end{algorithm}

\begin{figure}
	\begin{subfigure}[b]{0.45\linewidth}        
		\includegraphics[scale=0.35]{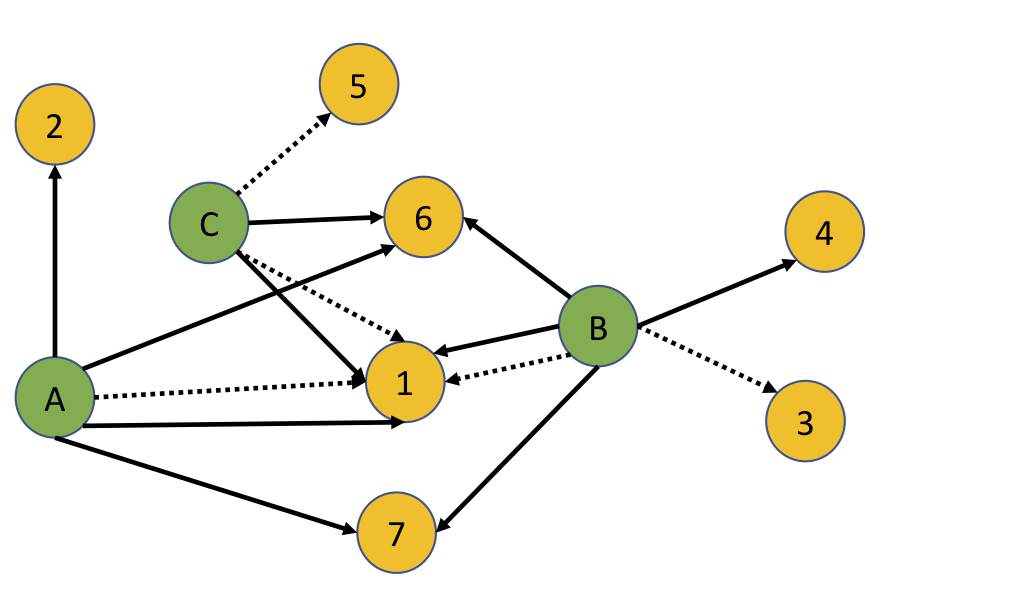}
		\caption{Original Subgraph}
		\label{fig:multi_ent_ori}
	\end{subfigure}
	\begin{subfigure}[b]{0.45\linewidth}        
		\includegraphics[scale=0.35]{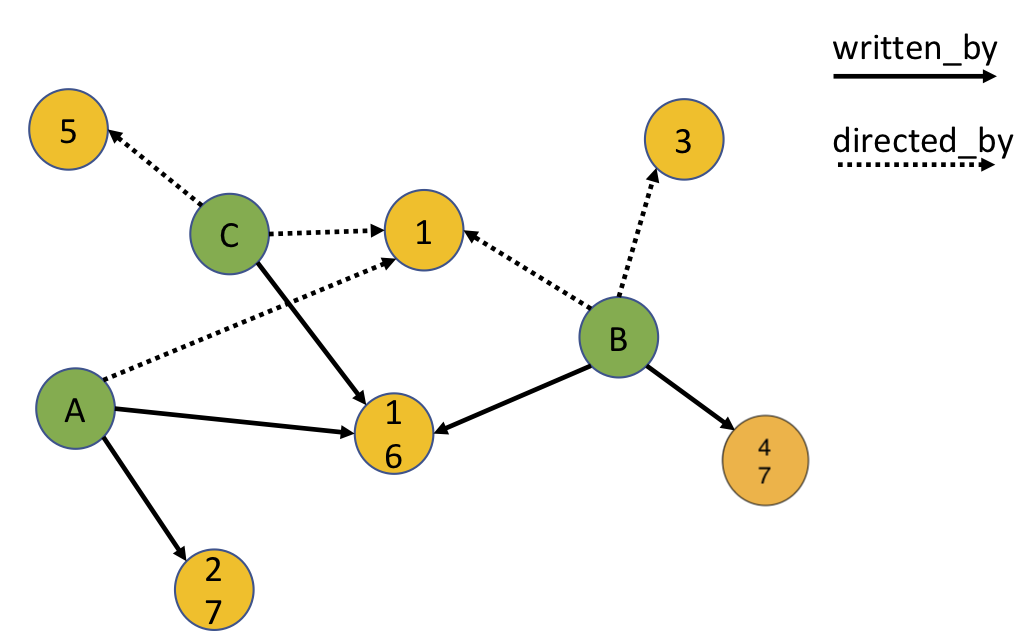}
		\caption{Grouped Subgraph}
		\label{fig:multi_ent_group}
	\end{subfigure}
	\caption{An example of grouping subgroup with three question entities: A, B and C; nodes in yellow are people and green nodes are movies}
	\label{fig:appendix}
\end{figure}

The algorithm for questions with multiple entities is based on relations and intersection sets, illustrated in Figure~\ref{fig:appendix}. It first categorizes neighbours of question entities based on relations. For example, it categorizes three neighbours (1, 5 and 6) of Entity C into ``written\_by(1, 6)" and ``directed\_by(1, 5)". Then, the algorithm finds the intersection node set for each relation. The intersection node set would be a supernode and the rest would be another supernodes. For example, the intersection node set for relation ``written\_by" is (1, 6) so Entity B has two neighbours linked by relation ``written\_by": (1, 6) and (4, 7). In this case, the summary graph in Figure~\ref{fig:multi_ent_group} is capable of answering questions ``who directed movie A, movie B and movie C?'' and ``who wrote movie A, movie B and movie C?'' with different entity representations. How GS-KGQA represents supernodes in the summary graph is introduced in the next section.
\subsection{Summary Graph Embedding Module}
Given a question and the summary graph, this module embeds each supernode $v_i^{(s)}$ in the summary graph to a fixed dimension vector. This is done by using a single-layer variant GCN with RCNN. RCNN uses a recurrent structure to capture contextual information and a max-pooling layer to capture the key components in texts. In our work, the RCNN's input is a question, and the outputs are the probabilities of the question belonging to all relations. 
The probabilities are then passed to the GCN as relational weights. As a result, the final representations of supernodes take the simple form:
\begin{equation}
h_i^1 =\frac{ h_i^0 + \sum_{j \in N_i^r} h_j^0 * w_r}{1+|N_i^r|} \label{eq:myRGCN}
\end{equation}
where $h_i^1$ represents $v_i^{(s)}$'s final representation; $h_i^0$ is $v_i^{(s)}$'s initialisation; $w_r$ is the relational weight from RCNN; $N_i^r$ contains all neighbours linked with relation $r$. $v_i^{(s)}$'s initialisation is:
\begin{equation}
    h_i^0 = \frac{\sum_{n_j \in v_i^{(s)}}t_j}{|v_i^{(s)}|}\label{eq:nodeEmb}
\end{equation}
where $n_j$ is the $j$-th member in the supernode $v_i^{(s)}$; $t_j$ is $n_j'$'s representation. To take additional semantic information into account, $t_j$ is the word vector.
\subsection{Answer Selection Module}
Given a question and the summary graph embedding, the answer selection module selects one supernode that best answers the question. This is achieved by estimating probabilities of a supernode $v_i^{(s)}$ given the question embedding:
\begin{equation}
answer = \underset{i}{max}(P(v_i^{(s)}| q)) = \underset{i}{max}(softmax(\textbf{H} \cdot LSTM(q)))
\end{equation}
where $q$ is the question, $\textbf{H}$ is a matrix whose each row represents the final representation $h_i^1$, $P(v_i^{(s)} | q)$ means the probability of supernode $v_i^{(s)}$ being an answer given the question $q$, symbol ``$\cdot$" is the dot product.
\section{Experiments}
We evaluated the effectiveness and robustness of the proposed GS-KGQA on widely adopted KGQA benchmarks. In this section, we aim to investigate the effectiveness of the proposed GS-KGQA when compared with other KGQA baselines. As the proposed GS-KGQA has two main areas: graph summarization algorithms and graph embedding, we also investigate the impact of these two proposals and how they impact the GS-KGQA system as a whole.
\subsection{Datasets}
We investigated four popular and publicly available KGQA datasets: WikiMovies\footnote{https://research.fb.com/downloads/babi/}, WebQuestionsSP\footnote{https://www.microsoft.com/en-us/download/details.aspx?id=52763}, WC2014\footnote{https://github.com/zmtkeke/IRN} and SimpleQuestions\footnote{https://github.com/davidgolub/SimpleQA/tree/master/datasets/SimpleQuestions}. SimpleQuestions was abandoned because all questions only have one answer, which cannot show the effectiveness of our graph summarization algorithms. For the other three datasets, there are over 60\% of questions with multiple answers. Also, entity linking for SimpleQuestions replies on the Freebase API, which is no longer available. 
We did not find any single-relation KGQA dataset containing questions with multiple entities, so we created a dataset called Two-Entity WikiMovies from the original WikiMovies dataset.
\begin{enumerate}
    \item \textbf{WikiMovies} is a popular dataset of single-relation KGQA with 96k questions. Its KG is from the movie domain, including 43k entities, nine predicates,
and 135k triples.
    \item \textbf{WebQuestionsSP} has a smaller scale of
questions but a larger scale of KG. It contains thousands of natural language questions based on Freebase\footnote{https://developers.google.com/freebase} , which has millions of entities and triples. Its questions are either single-relation or multi-relation. We used the single-relation questions.
    \item \textbf{WC2014} is from the football domain and based on the KG about football players participating in FIFA World Cup 2014. Its KG has 6k triples, 1k entities and ten predicates.
     \item \textbf{Two-Entity WikiMovies} was created from WikiMovies. We used regular expressions to learn the question patterns from WikiMovies. It contains 14k questions and uses the same KG as WikiMovies.
\end{enumerate}
\subsection{Setup}
GS-KGQA was implemented using Python, Pytorch and Neo4j. Neo4j is a Java-based open-source graph database. Neo4j provides query language, Cypher, to access data. All entities and predicates were initialised by self-trained word2vec vectors. We crawled the first paragraphs on Wikipedia for all entities. These paragraphs were used to train word vectors. We tuned the following hyperparameters with grid search:  (1) the hidden size for LSTM ([100,
150, 200]); (2) learning rate ([0.01, 0.001, 0.0001]); (2) the word dimension ([100, 150, 200]). As a result, we set hidden size=100, learning rate=0.01 for the Stochastic Gradient Descent (SGD) optimizer and dimension=150.
\subsection{Effectiveness of GS-KGQA}
We now answered the first research question asked at the beginning
of this section, i.e., how effective is GS-KGQA. We included six KGQA baselines:
\begin{itemize}
\item KV-Mem (2016)~\cite{miller2016key} used the key-value memory to store knowledge and conducted QA by iteratively reading the memory. This work released WikiMovies.
\item GRAFT-Nets (2018)~\cite{sun2018open} extracted a question-related subgraph from the entire KG with heuristics and then used variant GCN to predict the answer.
\item EmbedKGQA (2020)~\cite{saxena2020improving} took KGQA as a link prediction task and incorporated KGE with Bert to predict the answer.
\item SRN (2020)~\cite{qiu2020stepwise} was based on reinforcement learning. It leveraged beam search to reduce the number of candidates.
\item TransferNet (2021)~\cite{shi2021transfernet} computed activated scores for relations in relation analysis.
\item QA2MN (2021)~\cite{li2021question} dynamically focused on different parts of the questions in question analysis using a memory network. It used KGE to embed the KG.
\end{itemize}

As shown in the introduction above, all the baselines have taken
advantage of popular techniques, such as Bert, KGE, GCN and reinforcement learning, to advance their methods. For WikiMovies and WC2014, we used their results reported in the corresponding papers or the authors’ implementations. Note that we used single-relation questions in WebQuestions as introduced in Section 4.1; hence, we run the single-relation WebQuestions on their implementations for those that provided source codes publicly. The performances of the different methods are listed in Table~\ref{tab:result}.
\begin{table}
\caption{Performance of baselines on all datasets (hits@1)} \label{tab:result}
\centering
\begin{tabular}{c|c|p{2cm}|c|c}
\hline
\textbf{Model}  &\multicolumn{4}{c}{\textbf{Dataset}} \\
\hline
 & WikiMovies & Two-Entity WikiMovies & WebQuestionsSP&WC2014 \\
 \hline
KV-Mem &93.9 &\hfil  -& 31.3&87.0\\
GRAFT-Nets & 96.8&\hfil - &56.2&98.5 \\
EmbedKGQA & 97.5&\hfil -&80.7&98.2\\
SRN&97.0&\hfil -&-&\textbf{98.9}\\
TransferNet &97.5&\hfil -&-&-\\
QA2MN&-&\hfil -&-&98.6\\
GS-KGQA &\textbf{98.2}&     \textbf{\hfil 97.6}        &\textbf{84.6}&98.2\\
\hline
\end{tabular}
\end{table}

From the results in Table~\ref{tab:result}, we have three observations. First, the proposed system GS-KGQA outperforms all the baselines on WikiMovies and WebQuestionsSP. GS-KGQA achieves similar hits@1 with other baselines on WC2014. Second, most models performed badly on WebQuestionsSP because WebQuestionsSP has a relatively small number of training examples but uses a large KG. It makes the model training much harder, but GS-KGQA can still achieve a 3.9\% improvement when compared to EmbedKGQA. Third, the hits@1 on Two-Entity WikiMovies demonstrates that the proposed GS-KGQA could help the QA task on questions with multi-entities, which is hardly addressed by previous works. Therefore, no comparison can be performed to other baseline techniques.
\subsection{Impact of the Graph Summarization Algorithms}\label{sec:q1}
We now study how much could our proposed graph summarization algorithms contribute to the whole system. We found that all baselines tend to predict only one answer. It suggests that it is necessary to study questions with an uncertain number of answers. Our graph summarization algorithms were proposed to fill this gap.

We first validated the robustness of our graph summarization algorithms, i.e., if our graph summarization algorithms can group nodes correctly. Then, we wrote Cypher snippets to query the members of each supernode from the KG. Finally, we compared the members with the generated supernodes from our algorithms. All datasets achieve 100\% accuracy, which is because our algorithms can be treated as a set of reliable rules.

\begin{table}
\caption{This table shows the results (hits@1/recall) on questions with multiple answers} \label{tab:compare-multi}
\centering
\resizebox{0.99\textwidth}{!}{
\begin{tabular}{c|c|c|c|c}
\hline
\textbf{Model}  &\textbf{WikiMovies} & \textbf{WC2014} & \textbf{WebQuestionsSP} & \textbf{Two-Entity WikiMovies} \\
\hline
KV-Mem&95.1/24.5&88.1/19.7&31.0/6.2&-\\
GRAFT-Nets &97.2/26.0& 97.6/26.3 & 54.8/17.8&-\\
EmbedKGQA&96.2/25.0&97.9/27.1&83.5/39.6&-\\
GS-KGQA\_noGS &96.3/25.0&98.0/27.3&82.3/38.7&97.7/26.1\\
GS-KGQA &96.6/89.7& 98.4/98.5 & 81.1/87.1&97.6/98.7\\
\hline
\end{tabular}
}
\end{table}
We then conducted an ablation test and a comparison test to show the contribution of our graph summarization algorithms. We extracted questions with multiple answers (at least two answers) from all datasets and compared the hits@1 and the recall of actual answers on hits@1. The performance of GS-KGQA when not using graph summarization algorithms (denoted as GS-KGQA\_noGS) and the performance of baselines when testing on questions with multiple answers are shown in Table~\ref{tab:compare-multi}. From the results, we have two major observations. First, GS-KGQA\_noGS achieves similar hits@1 to GS-KGQA. It means that other modules in GS-KGQA are robust. Second, GS-KGQA has similar hits@1 performance with other baselines, but GS-KGQA significantly outperforms the recall. It indicates that the critical role of our graph summarization algorithms is to solve the multi-answer issue, which meets one of the objectives of this paper.

In summary, our graph summarization algorithms convert a multi-label multi-class classification problem to a unary label multi-class classification. The prediction of GS-KGQA may happen to be a supernode containing all answers. Our graph summarization algorithm's robustness and the greater recall boost help GS-KGQA to answer questions with an uncertain number of answers.
\subsection{Impact of the Graph Embedding}\label{sec:q2}
To study how much could the RCNN contribute to the proposed GS-KGQA and whether it can be done without the RCNN, we included an ablation test. The performance of GS-KGQA when not using the RCNN is denoted as GS-KGQA\_noRCNN. 
To further validate if the variant GCN helps in the graph embedding, we replaced GCN with three scalable KGE methods in the comparison: TransE~\cite{bordes2013translating}, TransH~\cite{wang2014knowledge} and ComplEx~\cite{trouillon2016complex}.

From the results in Table~\ref{tab:withourRCNN}, we have three major observations. First, GCN +RCNN beats KGE+RCNN in all datasets. Our variant GCN could achieve better performance than KGE because our variant GCN tends to aggregate information from direct neighbours. It is more suitable for the single-relation KGQA. Additionally, our variant GCN was initialised with semantic information. It helps GS-KGQA analyse answers based on both semantic and structural spaces; however, KGE is only in the structural space. Second, GS-KGQA\_noRCNN still outperforms KGE-based GS-KGQA in some datasets. For example, GS-KGQA\_noRCNN achieves 12\% improvement when GS-KGQA is based on ComplEx in Two-Entity WikiMovies. 
It further validates the robustness of our GCN. Third, RCNN indeed could improve the performance of GS-KGQA in all datasets. GS-KGQA achieves 16\%, 5.9\%, 7.4\%, 3.5\% higher hits@1 compared to GS-KGQA\_noRCNN.
\begin{table}
\caption{Comparison among variants; the statistics are hits@1}\label{tab:withourRCNN}
	\centering
	\resizebox{0.99\textwidth}{!}{
	\begin{tabular}{c|c|c|c|c}
\hline
\textbf{Model} & \multicolumn{4}{c}{\textbf{Dataset}}\\
\hline
&WikiMovies&WebQuestionsSP&WC2014&Two-Entity WikiMovies\\
\hline
GS-KGQA\_noRCNN &82.2 & 78.7 &90.8& 94.1\\
GS-KGQA& \textbf{98.2}&\textbf{84.6} & \textbf{98.2}&\textbf{97.6}\\
GS-KGQA\_TransE & 62.9&        55.4     & 62.7& 61.2\\
GS-KGQA\_TransH & 73.2 &       60.1     & 75.8& 74.2\\
GS-KGQA\_ComplEx &83.3&          65.4   & 86.9 & 82.1\\
 \hline
	\end{tabular}
	}
\end{table}

In summary, the great improvement of GS-KGQA compared to GS-KGQA \_noRCNN confirms that the relation analysis of the single-relation KGQA could be converted to the classification task. Moreover, assigning the classification probabilities to the GCN would enhance the embedding quality.
\section{Conclusion}
In this paper, a system GS-KGQA system has been proposed to answer single-relation questions. GS-KGQA combines GCN with RCNN to ensure that embeddings are propagated more along with edges relevant to the question. The RCNN converts the relation analysis of single-relation questions to the classification task. GS-KGQA utilizes graph summarization algorithms to tackle the issue that traditional KGQAs cannot answer questions with multiple entities and an uncertain number of answers. Experiment results showed GS-KGQA's effectiveness, which beats the hits@1 of all baselines by 0.7\% and 3.9\% on the best comparison baseline technique on WikiMovies and WebQuestionsSP, respectively. This paper also showed the importance of the use of the proposed graph summarization algorithms and the use of RCNN in the graph embedding in the proposed GS-KGQA.

\end{document}